\documentclass{article} 
\usepackage{PRIMEarxiv}

\usepackage{amsmath,amsfonts,bm}

\def\eqref#1{equation~\ref{#1}}

\def\1{\bm{1}}

\def\rvx{{\mathbf{x}}}

\def\mI{{\bm{I}}}

\DeclareMathAlphabet{\mathsfit}{\encodingdefault}{\sfdefault}{m}{sl}
\SetMathAlphabet{\mathsfit}{bold}{\encodingdefault}{\sfdefault}{bx}{n}

\def\gN{{\mathcal{N}}}

\usepackage{hyperref}
\usepackage{url}
\usepackage{graphicx}

\title{Targeted Image Reconstruction by Sampling Pre-trained Diffusion Model}

\author{Jiageng Zheng \\
Department of Computer Science\\
Lehigh University\\
\texttt{jiz322@lehigh.edu} \\
}

\newcommand{\bmu}{\boldsymbol{\mu}}
\newcommand{\bphi}{\boldsymbol{\phi}}
\newcommand{\bepsilon}{\boldsymbol{\epsilon}}

\newcommand{\bsigma}{\boldsymbol{\sigma}}
\newcommand{\bx}{\textbf{x}}
\newcommand{\qP}{q_{\bphi}}
\newcommand{\bz}{\textbf{z}}

\begin{document}

\maketitle

\begin{abstract}
A trained neural network model contains information on the training data. Given such a model, malicious parties can leverage the ”knowledge” in this model and design ways to print out any usable information. Therefore, it is valuable to explore the ways to conduct a such attack and demonstrate its severity. In this work, we proposed ways to generate a data point of the target class without prior knowledge of the exact target distribution by using a pre-trained diffusion model.
\end{abstract}

\section{Introduction}

Any machine learning model has learned something about the training data, so the leakage of a trained model may cause privacy concerns. If the training data contains private information, adversaries can leverage the visible information and capture the private data they want. Several studies have explored this possibility. For example, the membership inference attack (MIA) \cite{https://doi.org/10.48550/arxiv.2103.07853} compromises privacy by predicting whether a data point is a part of the model's training data. Another study proposed a reconstruction attack \cite{https://doi.org/10.48550/arxiv.2201.04845} which aims to reconstruct an exact data point in the training set. However, it requires the attacker to have more knowledge beyond the trained model, such as all the data points, neural networks' weight initialization, and learning rate schedule. 

This work focuses on the reconstruction attack as in \cite{https://doi.org/10.48550/arxiv.2201.04845}, but it requires less prior knowledge. More specifically, given a trained classification model and a distribution of the attack target, our adversary can generate the portrait of a previously invisible class (as shown in Figure 1). This type of reconstruction attack can also threaten the privacy of the training data. For example, imagine we have a face recognition model which classifies the image input as a specific person. If a malicious party retrieves this model, the party can generate a sample image for each person.

\begin{figure}[h]
\begin{center}
\includegraphics[scale=0.5]{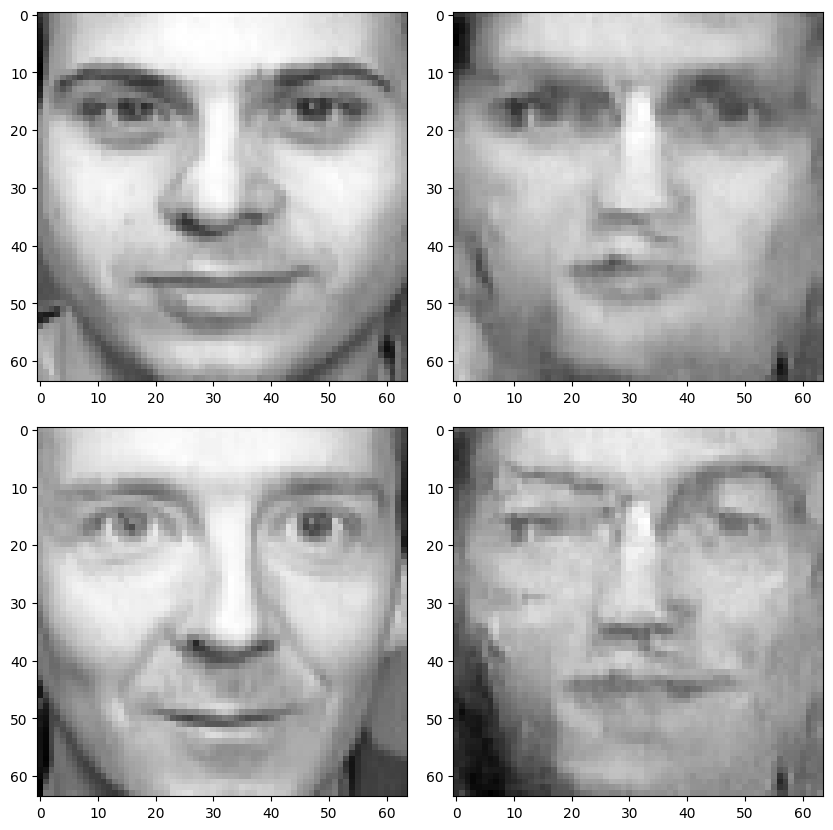}
\end{center}
\caption{The images in the left column are from the training set. The images in the right column are generated by the adversarial using the provided classification model and part of the visible training set. Notice that the visible training set does not contain the images on the left column nor other images of the same person, so the images in the left column are supposed to be private.}
\end{figure}

A human knows what his/her best friend looks like, but if this human is asked to draw his/her best friend a picture, this human may fail due to poor drawing skills. This phenomenon also applies to the trained neural network models. A trained classification model has learned how to process an image and do the classification. Suppose we query this model for a sample image of a specific class by fine-tuning the model's input. In that case, the tuned input will be a noise (Figure 2), regardless it grants $100\%$ classification confidence on the attack target. This happens because the classification model is not perfectly robust. Therefore, we use the distribution of attack targets to guide image generation.

\begin{figure}[h]
\begin{center}
\includegraphics[scale=0.5]{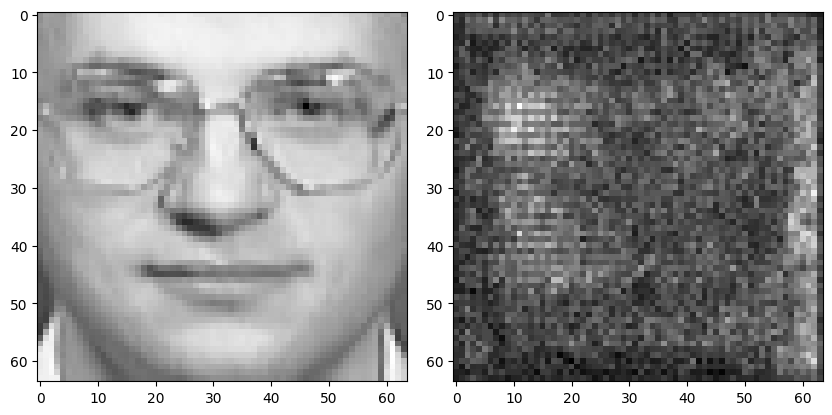}
\end{center}
\caption{This image on the right is generated by maximizing the classification possibility on an attack target (image on the left). The classifier classifies the adversarial noise to be $99\%$, the attack target. However, the generation of an adversarial noise does not satisfy our goal.}
\end{figure}

There are two requirements to evaluate a reconstructed image. First, this image should maximize the classification confidence of the attack target. Second, it should belong to the distribution of the attack target. Since the distribution of the attack target is invisible, we can approximate this using other distributions (for example, for a face recognition model, any distribution of face images can approximate the distribution of the attack target).

To satisfy the requirement of the generated distribution, we will use latent variable generative models (LVGM) to learn from a distribution. There are various types of LVGMs, and the options include the Generative Adversarial Network (GAN) \cite{https://doi.org/10.48550/arxiv.1406.2661}, the Variational Autoencoder (VAE) \cite{Kingma_2019}, and the diffusion related models \cite{https://doi.org/10.48550/arxiv.2006.11239}, \cite{https://doi.org/10.48550/arxiv.2106.06819}, and \cite{https://doi.org/10.48550/arxiv.2211.03264}. For different LVGMs, the methodologies for this reconstruction are also different, and the diffusion model is the focus of this work.

Notice that the goal of this work is different than \cite{https://doi.org/10.48550/arxiv.2201.04845}, where they aimed to reconstruct the exact image in the training set. However, here we want to reconstruct an image with the same class label and a similar distribution to the attack target. 

Moreover, the difference between this task and conditional image generation is that conditional image generation provides the entire data set, giving full knowledge of all classes. In contrast, for this work, a trained neural network model provides the knowledge of target classes in an implicit manner.

\section{Background}
\label{gen_inst}
\subsection{Conditional Generative Adversarial Networks}
In a conditional generative adversarial network \cite{https://doi.org/10.48550/arxiv.1611.07004}, we train a generator to achieve some goal and deceive the discriminator so that the discriminator cannot distinguish the target domain from the domain of the generated results.
The objective of the generator of conditional GAN can be expressed as
\begin{align}
    G^*  = \arg\min_G\max_D \mathcal{L}_{cGAN}(G,D) + \lambda \mathcal{L}_{L1}(G).\label{full_objective}
\end{align}
The first part of the expression showcases how the generator is trained to deceive the discriminator, and the second part represents the generator's goal. For our work, the generator's goal refers explicitly to maximizing the classification confidence on the attack targets. 

The conditional GAN learns from a distribution by evolving a generator and a discriminator. The generator is sufficiently trained when the generation result is sufficiently good or when the discriminator cannot be optimized.

\subsection{Variational Autoencoders}

The neural networks used by the variational autoencoders typically include two modules: an encoder which maps the input into deep latent space, and a decoder to generate result from latent features. An example of VAE encoder \cite{Kingma_2019} is expressed as below. Given the input $x$, the distribution of latent features is a normal distribution according to the mean and variance predicted by the encoder neural network.

\begin{align}
(\bmu, \log \bsigma) &= EncoderNeuralNet(\bx)\\
\qP(\bz|\bx) &= \mathcal{N}(\bz ; \bmu, \text{diag}(\bsigma))
\end{align}

The decoder neural network maps the distribution of latent features to the distribution of the inputs. The training of this decoder uses self-supervised learning (SSL). For example, the most straightforward training objective is whether the decoder can recover the original input image.

\subsection{Diffusion Models}
We train the diffusion model using the expression below \cite{https://doi.org/10.48550/arxiv.2006.11239}. The $\rvx^{(\alpha_i)}$ represents the noised input at the $\alpha_i$ step, and the neural network aims to predict the noise added.

\begin{align}
    \ell_{\text{diff}}(\rvx; w, \theta) := \sum_{i=1}^{T} w(\alpha_i) \textbf{E}_{\epsilon \sim \gN(0, \mI)}[\|{\epsilon - \epsilon_{\theta}(\rvx^{(\alpha_i)}, \alpha_i)}\|_2^2] , \quad \rvx^{(\alpha_i)} := \sqrt{\alpha_i} \rvx + \sqrt{1 - \alpha_i} \epsilon \label{eq:diffusion-training-obj}
\end{align}

Given an image of Gaussian noises, the diffusion model samples a denoised result by applying steps of the denoise process (Figure 3). Denoising is done by having a neural network predict the noise added on each step. A denoising step can be expressed below (algorithm 2 from \cite{https://doi.org/10.48550/arxiv.2006.11239}).

\begin{align}
\bx_{t-1} = \frac{1}{\sqrt{\alpha_t}}\left( \bx_t - \frac{1-\alpha_t}{\sqrt{1-\bar\alpha_t}} \bepsilon_\theta(\bx_t, t) \right) + \sigma_t \bz
\end{align}
In the expression (5), the $x_{t-1}$ represent the result after a step of denoising. In the most straightforward setup, the $\alpha$ and the $\sigma$ are fixed constants computed using the variance schedule. The $z$ is a noise in Gaussian distribution. The neural network takes a tuple of $x_t$ and $t$ and predicts the noise on step $t-1$.

\begin{figure}[h]
\begin{center}
\includegraphics[scale=0.2]{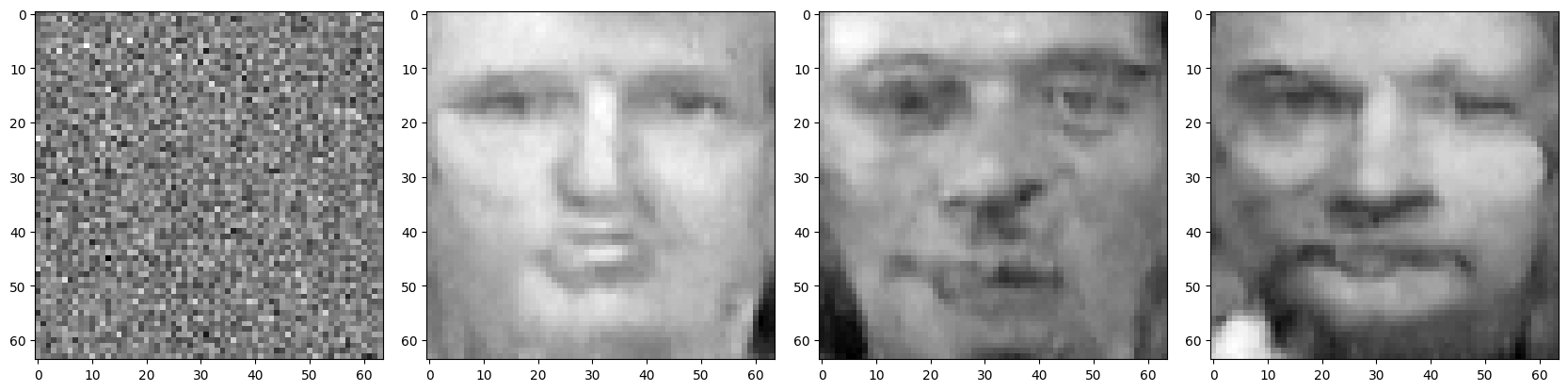}
\end{center}
\caption{The leftmost image is a fixed noise input into the sampling process of a trained diffusion model, and other images are the result of sampling. Even though we fixed the first input noise, the random Gaussian noises are involved in each step of the denoising, so the sampling results are different.}
\end{figure}

\section{Methodologies}
\label{headings}

\subsection{GAN Method}
We designed a method based on the conditional generative adversarial network \cite{https://doi.org/10.48550/arxiv.1611.07004}. The generator $G$, takes a Gaussian noise $\mathcal{N}(0,1)$ with the shape of the target image and outputs the same shape. The neural network architecture for the generator can be the U-net \cite{https://doi.org/10.48550/arxiv.1505.04597}. The discriminator $D$ takes an image as input and outputs the prediction of whether this image is fake (generated by the generator) or real (belonging to the target distribution).

We designed the loss function to train the generator for this reconstruction attack as below.
\begin{align}
      L_g = L_{CrossEntropy}(C(G(\mathcal{N}(0, 1))), y_{target}) + \alpha L_{CrossEntropy}(D(G(\mathcal{N}(0, 1))), 1)
\end{align}
The $C$ represents the classification model visible to adversaries; the $G$ is the generator; the $D$ is the discriminator. In the first part of this loss function, the generator's objective is to create an image, input this image to the classifier, and maximize the confidence that this generated image to be classified as the attack target. The image generated is input to the discriminator as in the second part of this loss function. The generator wants the discriminator to classify this image as "real" (or belongs to the target distribution). Notice that this objective function is similar to (1), while the only difference is that the generated image is specified to maximize the classification result of the attack target. 

To train this generator, we first generate a temporary data set for the discriminator, which includes both generated images and the images from the target distribution. Then, we train the discriminator, evolve the generator, and generate a new training set for the discriminator to start the new iteration until the generator is good enough.

\subsection{Sampling Pre-trained Diffusion Model}
For the image interpolation tasks, the VAE-based methods handle them by gradually adjusting the latent features of the decoder's input. In contrast, diffusion models handle them by gradually adjusting the intermediate noises. The well-trained VAE generator and diffusion model have learned how to generate an image to the target domain, so when interpolating two images, the generated interpolations will also belong to the target domain. For this reconstruction attack, the insight can be interpreted from the perspective of image interpolation: would there be an interpolation of images that happens to be our attack target?

For a well-trained diffusion model, we want to find a specific Gaussian noise from the distribution $\mathcal{N}(0,1)$. We sample the diffusion model using this noise and input the denoised image into the classifier, and we want this specific noise to maximize the prediction of the attack target.
The expression of the loss function is as below. Notice that we want to optimize the specific noise $x_t$. 
\begin{align}
      L = L_{CrossEntropy}(C(denoise(x_t)), y_{target})
\end{align}
The denoising process is the same as the formula (5). We fix the Gaussian noise $z$ for each denoising step so that we are only optimizing one path of denoising. The denoising process includes t steps. For each step, the previous noise $x_t$ contributes to a part of $x_{t-1}$'s computation, which is a suitable property since this is similar to the residual connections in the ResNet \cite{https://doi.org/10.48550/arxiv.1512.03385}, aiming to alleviate vanishing gradient issues.

Notice that if it is large, the computational graph for the gradient computation will also be significant. However, the modern computer with more than 256GB memory can still afford this computation.

\section{Experimental setup}
\label{others}

\subsection{Data set and Preparations}

The data set we used is the Olivetti Faces. It contains facial images of 40 different people. There are ten images for each person, so it has 400 images in total. For classification models, the task is to classify the input image to the specific person. We have seven images for training and three for each person for validation. We trained a simple 2-layer CNN model $C_2$, which gives $96.7\%$ top-1 validation accuracy for classifying 40 classes. We also trained a VGG11 model \cite{https://doi.org/10.48550/arxiv.1409.1556} $c_{vgg}$, intentionally made this vgg model overfits, but it still has top-1 validation accuracy for $90\%$. 

We set the attack targets for the reconstruction attack to range from person 1 to person 20. We assumed the adversarial obtains an estimation of target distribution by owning the rest of the training set (200 images from person 21 to person 40). 

For our GAN method, as discussed in 3.1, the architecture of the generator is U-net \cite{https://doi.org/10.48550/arxiv.1505.04597}, and the discriminator is a simple 2-layer CNN model. In each iteration, the training data for the discriminator contains the images of person 21 to person 40 (200 images) and another 200 images generated by the current generator. We train the generator until the progress of generated result becomes unobservable.

We also trained a diffusion model on the images of the last 20 people. The neural network architecture to predict the noise is also U-net. We set the number of steps for adding or removing noise to 600, and the variance schedule ranges from $10^{-4}$ to $0.02$. 

Moreover, we built a variational autoencoder to make more comparisons. We conduct the image reconstruction for the VAE method by fine-tuning the features in latent space.

\subsection{Evaluation Metrics}
Different from the reconstruction attack in \cite{https://doi.org/10.48550/arxiv.2201.04845}, we cannot directly compute the difference between the generated result and the training set because we are not reconstructing any specific image but a class. One of the most straightforward ways is to conduct a human evaluation. For example, we can purchase the Amazon Mechanical Turk service (we did not do it).

A better way to evaluate our result is using similar evaluation metrics as conditional image generation. The evaluation metrics for conditional image generation on Cifar benchmarks use a pre-trained Inception v3 model \cite{https://doi.org/10.48550/arxiv.1512.00567}. One common evaluation metric is Frechet Inception Distance, where the Inception model extracts latent representations of the generated image. If we get a short distance between the extracted latent representation and that of the ground truth images, the generation is sufficiently good. However, we cannot use the FID metrics since the Inception model is not for facial images. To compromise, we use classification model $C_2$ to evaluate, which is a more robust model than the classifier $C_{vgg}$ we used to attack. For simplicity, we directly input the generated image into the classifier$C_2$ and collect the prediction confidence on the attack target.

\section{Experiment Results}

\subsection{Comparison of Different Ways of Image Generation}

\begin{figure}[h]
\begin{center}
\includegraphics[scale=0.2]{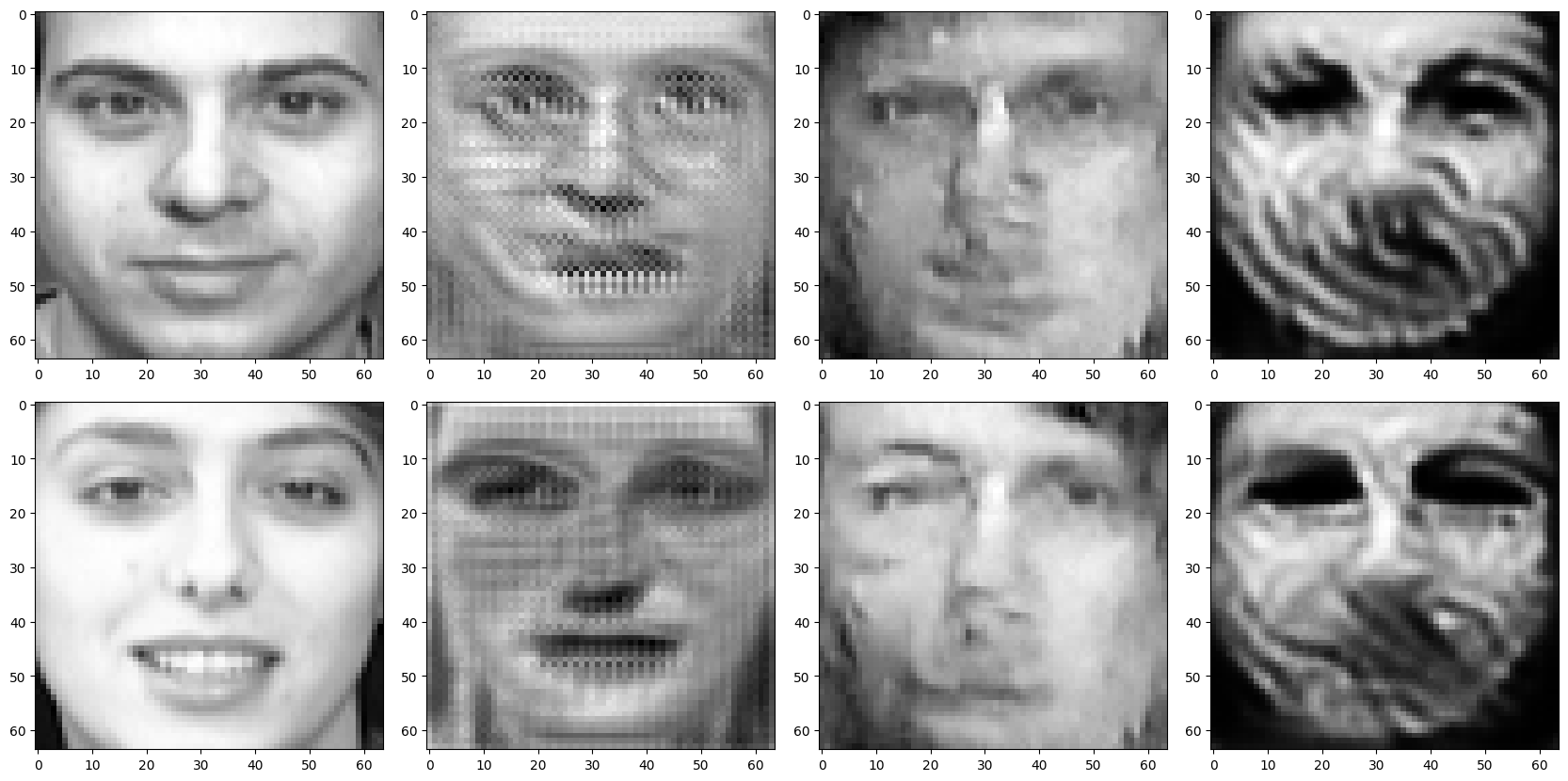}
\end{center}
\caption{From the left-most column to the right-most column, the images represent examples of attack targets, reconstruction results of GAN, reconstruction results of Diffusion, and reconstruction results of VAE. The first row is the attack's first target (the eighth person in the data set). The second row is the second attack target (the seventh person in the data set).}
\end{figure}

From Figure 4, we can get the qualitative result that the diffusion-based method can generate better human faces. Notice that all the methods used the $C_{vgg}$ classifier. To quantitatively analyze the generation of the target class, we input the generated images into the $C_2$ classifier and collect the confidence that the generated images are classified as the attack target. We show the results of the quantitative analysis in table 1. Using the Diffusion method to attack target2 gives the only satisfiable result. On the one hand, this result is reasonable since the $C_{vgg}$ is a classification model, it does not have complete knowledge of all the classes, and one successful case indicates the threat of this type of attack is indeed present. On the other hand, there is room to improve the attack methodologies. 

\begin{table}[t]
\caption{Quantitative Analysis of Generation Results}
\label{sample-table}
\begin{center}
\begin{tabular}{ll}
\multicolumn{1}{c}{\bf Confidence of being Predicted as Attack target}  &\multicolumn{1}{c}{\bf GAN$\setminus$VAE$\setminus$Diffuison}
\\ \hline \\
Target1        &$0.0003\setminus0.0002\setminus0.0000$ \\
Target2             &$0.0030\setminus0.1069\setminus\bf0.9077$ \\
\end{tabular}
\end{center}
\end{table}

\subsection{Effect of Different Learning Rates for Fine-tuning $x_t$}
When fine-tuning the noise, the loss value will always converge no matter how significant the learning rate is (we tried a range from 1e-5 to 1e10). However, when selecting different learning rates, the result of this diffusion method will be different (Figure 5). To explain this observation, we need to trace the gradient update during the back propagation and do more experiments.

\begin{figure}[h]
\begin{center}
\includegraphics[scale=0.2]{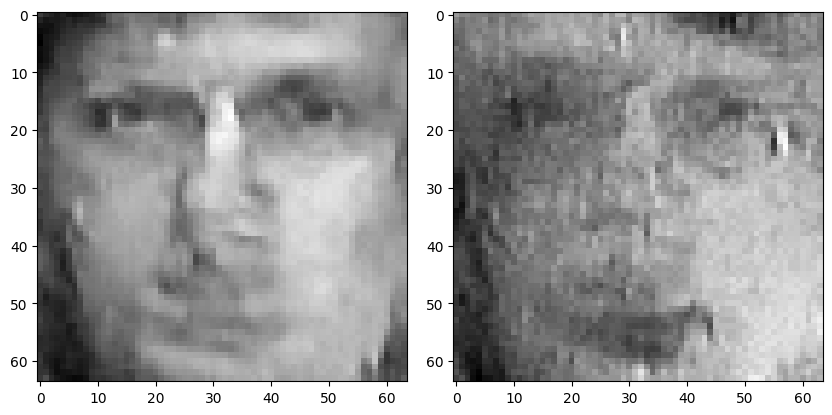}
\end{center}
\caption{Both generated image has a loss value of around $1.3$, but for the left image, we set the learning rate to $1$ and optimized for 16 iterations. In contrast, we set the learning rate for the right image to $10$ and optimized it for 9 iterations.}
\end{figure}

\section{Conclusions and Future Works}

In this work, we explored the potential approaches to generate an image for a target class when provided with a trained classification model and an approximate distribution of the attack target. The method of sampling a pre-trained diffusion model has great potential and deserves further research investment. Limited by time and our computational power, more thorough analysis and experiments should be done on the method we proposed in 3.2. Moreover, some observations need to be better explained, such as the effect of the learning rate discussed in 5.2. It is also a research direction to explore the relationship between the generation quality and the provided classification models.

There are new variants of the Diffusion model's sampling methods, for example, the denoising diffusion implicit models (DDIMs) \cite{https://doi.org/10.48550/arxiv.2010.02502}, which can give high-quality samples much faster. The diffusion model's sampling method improvements can significantly reduce this work's memory consumption since the gradient computation graphs dominate our memory consumption for the massive number of denoising steps. Therefore, it is also valuable to update the methodologies proposed in this work along with the quick evolution of the existing diffusion model's methodologies.

\bibliographystyle{unsrt} 
\bibliography{main}

\begin{thebibliography}{10}

\bibitem{https://doi.org/10.48550/arxiv.2103.07853}
Hongsheng Hu, Zoran Salcic, Lichao Sun, Gillian Dobbie, Philip~S. Yu, and Xuyun
  Zhang.
\newblock Membership inference attacks on machine learning: A survey, 2021.

\bibitem{https://doi.org/10.48550/arxiv.2201.04845}
Borja Balle, Giovanni Cherubin, and Jamie Hayes.
\newblock Reconstructing training data with informed adversaries, 2022.

\bibitem{https://doi.org/10.48550/arxiv.1406.2661}
Ian~J. Goodfellow, Jean Pouget-Abadie, Mehdi Mirza, Bing Xu, David
  Warde-Farley, Sherjil Ozair, Aaron Courville, and Yoshua Bengio.
\newblock Generative adversarial networks, 2014.

\bibitem{Kingma_2019}
Diederik~P. Kingma and Max Welling.
\newblock An introduction to variational autoencoders.
\newblock {\em Foundations and Trends{\textregistered} in Machine Learning},
  12(4):307--392, 2019.

\bibitem{https://doi.org/10.48550/arxiv.2006.11239}
Jonathan Ho, Ajay Jain, and Pieter Abbeel.
\newblock Denoising diffusion probabilistic models, 2020.

\bibitem{https://doi.org/10.48550/arxiv.2106.06819}
Abhishek Sinha, Jiaming Song, Chenlin Meng, and Stefano Ermon.
\newblock D2c: Diffusion-denoising models for few-shot conditional generation,
  2021.

\bibitem{https://doi.org/10.48550/arxiv.2211.03264}
Jingyuan Zhu, Huimin Ma, Jiansheng Chen, and Jian Yuan.
\newblock Few-shot image generation with diffusion models, 2022.

\bibitem{https://doi.org/10.48550/arxiv.1611.07004}
Phillip Isola, Jun-Yan Zhu, Tinghui Zhou, and Alexei~A. Efros.
\newblock Image-to-image translation with conditional adversarial networks,
  2016.

\bibitem{https://doi.org/10.48550/arxiv.1505.04597}
Olaf Ronneberger, Philipp Fischer, and Thomas Brox.
\newblock U-net: Convolutional networks for biomedical image segmentation,
  2015.

\bibitem{https://doi.org/10.48550/arxiv.1512.03385}
Kaiming He, Xiangyu Zhang, Shaoqing Ren, and Jian Sun.
\newblock Deep residual learning for image recognition, 2015.

\bibitem{https://doi.org/10.48550/arxiv.1409.1556}
Karen Simonyan and Andrew Zisserman.
\newblock Very deep convolutional networks for large-scale image recognition,
  2014.

\bibitem{https://doi.org/10.48550/arxiv.1512.00567}
Christian Szegedy, Vincent Vanhoucke, Sergey Ioffe, Jonathon Shlens, and
  Zbigniew Wojna.
\newblock Rethinking the inception architecture for computer vision, 2015.

\bibitem{https://doi.org/10.48550/arxiv.2010.02502}
Jiaming Song, Chenlin Meng, and Stefano Ermon.
\newblock Denoising diffusion implicit models, 2020.

\end{thebibliography}

\end{document}